\documentclass{article}

\PassOptionsToPackage{numbers}{natbib}
\usepackage[preprint]{neurips_2021}

\usepackage[utf8]{inputenc} % allow utf-8 input
\usepackage[T1]{fontenc}    % use 8-bit T1 fonts
\usepackage[breaklinks=true,colorlinks,bookmarks=false]{hyperref}       % hyperlinks
\usepackage{url}            % simple URL typesetting
\usepackage{booktabs}       % professional-quality tables
\usepackage{amsfonts}       % blackboard math symbols
\usepackage{nicefrac}       % compact symbols for 1/2, etc.
\usepackage{microtype}      % microtypography
\usepackage{xcolor}         % colors

\usepackage{makecell}
\usepackage{algorithm}
\usepackage{algorithmic}
\usepackage{graphbox}
\usepackage{graphicx}
\usepackage{bm}
\usepackage[export]{adjustbox}
\usepackage{natbib} 
\usepackage{amsmath}
\usepackage{subfigure}
\usepackage{tikz}
\usetikzlibrary{matrix,shapes,arrows,positioning,chains,calc}

\title{Automated Decision-based Adversarial Attacks}

\newcommand{\sysname}{AutoDA}

\author{%
  Qi-An Fu, Yinpeng Dong, Hang Su, Jun Zhu \\
  Dept. of Comp. Sci. \& Tech., Institute for AI, BNRist Center \\
  \texttt{\{fqa19,dyp17\}@mails.tsinghua.edu.cn, \{suhangss,dcszj\}@mail.tsinghua.edu.cn}
}

\begin{document}

\maketitle

\begin{abstract}
Deep learning models are vulnerable to adversarial examples, which can fool a target classifier by imposing imperceptible perturbations onto natural examples. In this work, we consider the practical and challenging decision-based black-box adversarial setting, where the attacker can only acquire the final classification labels by querying the target model without access to the model's details. Under this setting, existing works often rely on heuristics and exhibit unsatisfactory performance. To better understand the rationality of these heuristics and the limitations of existing methods, we propose to automatically discover decision-based adversarial attack algorithms. In our approach, we construct a search space using basic mathematical operations as building blocks and develop a random search algorithm to efficiently explore this space by incorporating several pruning techniques and intuitive priors inspired by program synthesis works. Although we use a small and fast model to efficiently evaluate attack algorithms during the search, extensive experiments demonstrate that the discovered algorithms are simple yet query-efficient when transferred to larger normal and defensive models on the CIFAR-10 and ImageNet datasets. They achieve comparable or better performance than the state-of-the-art decision-based attack methods consistently.
\end{abstract}

\section{Introduction}
\label{sec:introduction}

Deep learning models have demonstrated impressive performance on various pattern recognition tasks \cite{Krizhevsky:2012a0,He:201604,Devlin:2019ad,Graves:201383}. However, these models are vulnerable to adversarial examples \cite{Szegedy:2014cd,Goodfellow:2015c5}, which are maliciously crafted by adding small adversarial perturbations to natural images but can fool the target model. A number of adversarial attack methods have been developed \cite{Goodfellow:2015c5,Papernot:2016c4,Ilyas:20185d,Brendel:201866} to generate adversarial perturbations under various threat models, which help to identify the vulnerabilities of deep learning models and serve as a surrogate to evaluate adversarial robustness \cite{Carlini:20174b,Dong:20200a}.

With the rapid development of adversarial attack and defense methods, it is of great importance to evaluate the existing methods correctly and reliably \cite{Carlini:2019ba,Dong:20200a,Croce:202099}. It sometimes needs carefully designed adaptive attacks to evaluate the worst-case robustness of a particular defense in case of gradient obfuscation \cite{Athalye:201869,Tramer:20209b}. Those attack methods were manually designed by experts case by case, which requires considerable trial-and-error efforts. One may hope to automatically discover attack methods to reduce this burden, which can not only serve as a reasonable baseline for measuring the strength of human designed attack methods but also examine the rationality of assumptions made in these attack methods.

%\junz{add a sentence to have smooth connection; the other part of this paragraph needs polishing to reflect the change}
Such a desire to automate the attacks becomes even urgent when we consider the practical yet challenging setting of decision-based black-box attacks, % \cite{Brendel:201866}, 
%The decision-based black-box attacks are developed under the most challenging threat model, 
where the attacker can only query the target model for the final classification labels. 
%This threat model is yet more practical in real-world scenarios \cite{Brendel:201866}. 
Although various decision-based attack methods have been proposed \cite{Brendel:201866,Cheng:20191b,Dong:2019ae,Brunner:2019b4,Cheng:20200c}, many of them are much more heuristic and exhibit unsatisfactory performance, % many of them exhibit unsatisfactory performance due to the heuristic design, fu's comment: methods based on non heuristics optimization methods' performance is unsatisfactory too.
as compared to gradient-based white-box attack methods which could be optimal in some sense \cite{Madry:2018e9}.
%However, unlike gradient-based white-box attack methods which could be optimal in some sense \cite{Madry:2018e9}, many of existing decision-based black-box attack methods \cite{Brendel:201866,Dong:2019ae,Brunner:2019b4} are much more heuristic and exhibit unsatisfactory performance, 
Such a gap urges the need for automated attack methods more than other threat models, to understand the rationality of these heuristics, and even to discover new methods with better performance.

The problem of automatically discovering adversarial attack algorithms falls into the general direction of program synthesis,
% The automatic generation of adversarial attacks falls into the general direction of program synthesis,
which aims to automatically discover a program satisfying 
%the user intent expressed in the form of some speciﬁcation.
%solve an extremely hard problem of automatically discovering algorithms satisfying
a user intent specification \cite{gulwani2017program}. Many generic methodologies and techniques have been developed for program synthesis, 
%while domain knowledge is still needed to solve a specific problem in different domains. Previous program synthesis 
with the majority 
%works mainly 
focusing on software problems \cite{SolarLezama:200697,Gulwani:2011eb,Balog:2017ba}.
One the other side, the task of automating the process of solving machine learning problems is known as automated machine learning (AutoML) \cite{Feurer:2015ac}. 
%instead of machine learning problems. Besides, automated machine learning (AutoML) also aims to automate the process of solving machine learning problems \cite{Feurer:2015ac}, 
One most attractive direction of AutoML is neural architecture search (NAS), which aims to automatically discover good architectures of deep networks \cite{Zoph:201713}, while existing methods often start with expert designed layers. AutoML-Zero \cite{Real:202082} moves one-step further and shows the promise to search for a complete classification algorithm (e.g., two-layer neural networks) from scratch using basic mathematical operations as building blocks with minimal human participation.
However, the discovered classification algorithms are still far behind the current practice.

In this work, we propose to solve a practical yet challenging problem of decision-based adversarial attack with competitive performance by automatically searching for % simple yet query-efficient
attack algorithms.
We call our approach \textbf{Auto}mated \textbf{D}ecision-based \textbf{A}ttacks (\textbf{\sysname{}}).
%which is so-called \textbf{\sysname{}} (\textbf{Auto}mated \textbf{D}ecision-based \textbf{A}ttacks).
Technically, we design a search space constructed from basic mathematical operations, which provides sufficient expressiveness for the decision-based attack problem with affordable complexity. Similar search spaces are used by many program synthesis works aiming to automatically solve software problems \cite{Gulwani:2011eb}. Thus we adapt useful methodologies and techniques from program synthesis to \sysname{}. For example, we use an algorithm template \cite{Srivastava:201354} to alleviate the difficulty of the search problem and use pruning techniques based on logical constraints \cite{gulwani2017program} imposed by the algorithm template and the adversarial attack problem. The design choice of constructing search space from basic mathematical operations is also similar to the previous mentioned AutoML-Zero \cite{Real:202082}. 
However, due to the theoretical and practical differences between our problem and AutoML-Zero's, \sysname{} settles on quite different design choices and implementations, e.g., we use the static single assignment (SSA) form instead of the three-address code (TAC) form
% \hangx{full name for the first time, SSA and TAC}
used in AutoML-Zero to define the search space for better sample efficiency and computational performance in our use case of generating random programs, as detailed in Section~\ref{ssec:search_space}.

To explore this search space efficiently, we develop a random search algorithm combined with several pruning techniques and intuitive priors. To further reduce computational cost, we utilize a small and fast model for evaluating attack algorithms during the search. 
%Our discovered top performing algorithms on this fast model are simple. They are yet query-efficient when transferred to larger models. They share common operation sequence with existing attack methods, which illustrates the rationality of some heuristics in existing works. 
Despite the simplicity of the discovered top performing algorithms on this small model, they are also query-efficient when transferred to larger models and share common operation sequence with existing attack methods, which illustrates the rationality of some heuristics in existing works.
Our discovered algorithms consistently demonstrate comparable or better performance than the state-of-the-art decision-based methods when attacking normal and defensive models on the CIFAR-10 \cite{krizhevsky2009learning} and ImageNet \cite{Deng:2009ff} datasets.

\section{Related Work}
\label{sec:related_work}

\textbf{Adversarial attacks and defenses.} After deep learning models have been found to be vulnerable to adversarial examples \cite{Szegedy:2014cd,Goodfellow:2015c5}, lots of attack methods under different threat models have been developed recently. 
%Distinguished by the adversary's goal, we have the untargeted threat model where the attacker aims to cause misclassification of the target classifier model, and the targeted threat model where the attacker aims to cause misclassification to a adversary-desired target class. Distinguished by the distance metrics, we have the \(\ell_p (p = 2, \infty, \dots)\) norm threat models, where the attacker intents to minimized the magnitude of the adversarial perturbations measured by \(\ell_p\) norm. 
In general, existing attacks can be categorized into the white-box and black-box attacks. Under the white-box setting, the attacker has full knowledge about the target model, and thus various gradient-based attack methods can be applied, such as the fast gradient sign method (FGSM) \cite{Goodfellow:2015c5}, the projected gradient descent (PGD) method \cite{Madry:2018e9}, and the C\&W method \cite{Carlini:20174b}. In contrast, under the black-box threat model, the attacker has limited access to the target model. For example, under the score-based black-box threat model, the attacker can only acquire the output probabilities of the black-box model with a limited number of queries \cite{Ilyas:20185d,Uesato:2018e9,cheng2019improving}. The decision-based black-box setting is more challenging because the attacker can only obtain the final classification labels by querying the target model \cite{Brendel:201866,Dong:2019ae,Brunner:2019b4,Cheng:20191b,Cheng:20200c}. This setting is yet more practical in real-world scenarios \cite{Brendel:201866}.
Due to the security threat, various defense methods have been proposed to defend against adversarial attacks \cite{Madry:2018e9}. However, many of them cause obfuscated gradients and can be broken by adaptive attacks \cite{Athalye:201869}. Currently, the most effective defense methods are based on adversarial training \cite{Dong:20200a,Madry:2018e9}.

\textbf{Program synthesis.} Our approach also relates to program synthesis, whose core problem is to generate a program that meets an intent specification \cite{gulwani2017program}. Many program synthesis works use traditional techniques, e.g., SKETCH \cite{SolarLezama:200697} solves the programming by sketching problem using SAT solver and Brahma \cite{Gulwani:2011eb} can efficiently discover highly nontrivial up to 20 lines loop-free bitvector programs from basic bitvector operations using SMT solver. Recent works may use machine learning techniques, e.g., DeepCoder \cite{Balog:2017ba} solves the programming by example problem using neural-guided search. These works mainly focus on software problems instead of machine learning problems.
Static analysis techniques are essential in these works and they are also useful for our \sysname{}, as detailed below.

\section{Methods}
\label{sec:methods}

In this section, we present \sysname{} in detail. For simplicity, we particularly focus on untargeted attacks in this work, where the attacker aims to cause misclassification on the victim classifier.
%with the purpose to mislead the victim classifier.
Nevertheless, our approach can be extended to targeted attacks straightforwardly.

\subsection{Overview}
\label{ssec:overview}

Discovering an algorithm that satisfies an intent specification is an undecidable problem in general \cite{gulwani2017program}, and thus is extremely hard. One approach to reduce the difficulty of this problem is to provide a template for the algorithm, which reduces the problem down to searching for missing components in this template \cite{Srivastava:201354}. Inspired by this approach, we choose the random walk framework for decision-based attacks under the \(\ell_2\) norm as our algorithm template. This framework is first proposed by the Boundary attack \cite{Brendel:201866} and used by many later decision-based attacks \cite{Dong:2019ae,Brunner:2019b4}. 

% \yinpeng{Motivated by this principle} principle is a too-big word?

As outlined in Alg.~\ref{alg:framework}, the random walk process starts at an adversarial starting point \(\bm{x}_1\), which could be obtained by keeping adding different large random noises to the original example \(\bm{x}_0\) until finding one that causes misclassification. In each iteration of the random walk, the attacker executes the \texttt{generate()} function to generate the next random point \(\bm{x}'\) to walk to based on the original example \(\bm{x}_0\) and the best adversarial example \(\bm{x}\) already found. \(\bm{x}'\) is usually generated by applying transformations to a Gaussian noise. If \(\bm{x}'\) is adversarial and is closer to \(\bm{x}_0\) than the old adversarial example \(\bm{x}\), we update the adversarial example \(\bm{x}\) to \(\bm{x}'\) since we found a better adversarial example with a smaller perturbation.
There are also some hyperparameters inside the \texttt{generate()} function controlling the step size of the random walk process. After each iteration, the framework would collect the success rate of whether \(\bm{x}'\) is adversarial and adjust the hyperparameters according to the success rate of several past trials.

There are two missing components in this template --- the \texttt{generate()} function and the hyperparameter adjustment strategy.
%It is noted that there exist some hyperparameters inside the \texttt{generate()} function to control the step size of the random walk process.
The main difference between existing attack methods lies in their \texttt{generate()} functions, while their strategies for hyperparameter adjustment are similar, i.e., they all adjust their hyperparameters to make the step size smaller when the success rate is too low and vice versa. Without loss of generality, we only search for the \texttt{generate()} function to make our problem easier, and settle on a predefined negative feedback strategy for adjusting hyperparameters similar to existing works, as detailed in the supplementary material.
% A detailed description is provided in the supplementary material.

To solve our problem, we follow the generic methodology from program synthesis: define a search space for the \texttt{generate()} function, design a search method, and search for programs with top performance under some designed evaluation metrics. Before diving into the details, we provide an overview of \sysname{} first: (1) We choose a generic search space constructed from basic scalar and vector mathematical operations which provides sufficient expressiveness for our problem; (2) We use random search combined with several pruning techniques and intuitive priors to efficiently explore the search space; (3) We evaluate programs with a small and fast model on a small number of samples to reduce computational cost.
The whole system of \sysname{} is complex.
We will elaborate design choices as well as important implementation details for the rest of this section and include more implementation details in the supplementary material.

\begin{algorithm}[tb]
  \caption{Random walk framework for decision-based attacks under the \(\ell_2\) norm.}
  \label{alg:framework}
\begin{algorithmic}
  \STATE {\bfseries Data:} original example \(\bm{x}_0\), adversarial starting point \(\bm{x}_1\);
  \STATE {\bfseries Result:} adversarial example \(\bm{x}\) such that the distortion \(\| \bm{x} - \bm{x}_0 \|_2\) is minimized;
    
  \STATE \(\bm{x} \gets \bm{x}_1\); \(d_{\min} \gets \| \bm{x} - \bm{x}_0 \|_2\);
  \WHILE {query budget is not reached}
    \STATE \(\bm{x}' \gets \texttt{generate(}\bm{x}, \bm{x}_0\texttt{)} \);
    \IF{\(\bm{x}'\) is adversarial \AND \(\|\bm{x}'-\bm{x}_0\|_2 < d_{\min}\)}
      \STATE \(\bm{x} \gets \bm{x}'\); \(d_{\min} \gets \| \bm{x} - \bm{x}_0 \|_2\);
    \ENDIF
    \STATE update the success rate of whether \(\bm{x}'\) is adversarial;
    \STATE adjust hyperparameters according to the success rate;
  \ENDWHILE
\end{algorithmic}
\end{algorithm}

\subsection{Search Space}
\label{ssec:search_space}

Designing a search space is the art of trading off between expressiveness and complexity \cite{gulwani2017program}. On one hand, the search space should be expressive enough to include useful programs for the target problem. On the other hand, great expressiveness does not come for free --- it usually leads to high complexity. Searching over a complex space is both time-consuming and hard to implement. Instead of using a full-featured programming language like Python which provides more-than-needed expressiveness with high complexity, we choose to design a domain specific language (DSL) specialized for our problem that provides sufficient expressiveness with relative low complexity.

We list all the available operations in our \sysname{} DSL in Table~\ref{tab:ops_list}. These operations are basic mathematical operations for scalars and vectors, and all vector operations have geometric meaning in the Euclidean space. Then we construct our search space for the \texttt{generate()} function as all valid static single assignment (SSA) form programs in this DSL with a given length and a given number of hyperparameters.
%We choose the SSA form which is widely used in modern compilers \cite{Lattner:20045d} to define our search space instead of the three-address code (TAC) form used in AutoML-Zero \cite{Real:202082}. Though the SSA and TAC forms are equivalent in the sense that they can be converted to each other, the SSA form has better sample efficiency and computational performance than the TAC form in our use case of generating random programs.
In our use case of generating random programs, we choose the SSA form widely used in modern compilers \cite{Lattner:20045d} instead of the three-address code (TAC) form used in AutoML-Zero \cite{Real:202082} for better sample efficiency and computational performance. Although the SSA and TAC forms are equivalent in the sense that they can be converted to each other, when generating random programs in the SSA form, we can enforce many wanted properties of these programs explicitly and straightforwardly, e.g., limiting the number of hyperparameters and avoiding unused inputs and operations. In contrast, for the TAC form, we need to generate programs first, then check their properties and reject the failed ones. Moreover, checking a TAC form program requires almost as much work as converting it into an equivalent SSA form program. Consequently, this generate-then-check process hurts sample efficiency and computational performance. It is worth noting that the \sysname{} DSL only requires all vector variables to have the same dimension but does not restrict them to a specific dimension. This property of our DSL preserves the possibility for transferring the discovered programs to other datasets with different image shapes without modification, though hyperparameter's initial values might need extra tuning after changing the input dimension.

% \junz{this is a good point to differ from AutoML-zero, should highlight in Intro?} \fu{too detailed for intro, ``our problem practically differ from AutoML-zero, thus we settle on quite different design choices and implementation, e.g. we use the SSA form instead of the TAC form as detailed in Section~xxx?''}

We design the program to accept three parameters: \(\bm{x}\) and \(\bm{x}_0\) as in the \mbox{\texttt{generate(}\(\bm{x}, \bm{x}_0\)\texttt{)}} function from Alg.~\ref{alg:framework}, as well as a random noise \(\bm{n}\) sampled from the standard Gaussian distribution \(\mathcal{N}(\bm{0}, \mathrm{I})\). Instead of providing operations for generating random noise in the \sysname{} DSL, we provide the random noise as a parameter \(\bm{n}\). Combining with the SSA form, the program itself would be pure and more handy to do property testing efficiently.

The above designed \sysname{} DSL has sufficient expressiveness for the decision-based adversarial attack under the \(\ell_2\) norm problem. For example, we can implement the Boundary attack's \texttt{generate()} function with our \sysname{} DSL. We provide one possible implementation of it in the supplementary material. On the other hand, the \sysname{} DSL does not have high complexity since it has no control flow and has only ten unary and binary operations. However, this search space is still huge, because its size grows at least exponentially with the length of the program. As a result, we need to design and implement an efficient search method.

\begin{table}[t]
\caption{List of available operations in the \sysname{} DSL. The suffix of each operation's notation indicates the parameters' type of the operation, where \texttt{S} denotes scalar type, and \texttt{V} denotes vector type. For example, the \texttt{VS} suffix means the operation's first parameter is a scalar and second parameter is a vector. Detailed definitions are provided in the supplementary material.}
\label{tab:ops_list}
\vskip 0.1in
\begin{center}
\setlength{\tabcolsep}{12.0pt}
\begin{tabular}{lll}
  \toprule
  ID & Notation        & Description                            \\
  \midrule
  1  & \texttt{ADD.SS} & scalar-scalar addition                 \\
  2  & \texttt{SUB.SS} & scalar-scalar subtraction              \\
  3  & \texttt{MUL.SS} & scalar-scalar multiplication           \\
  4  & \texttt{DIV.SS} & scalar-scalar division                 \\
  5  & \texttt{ADD.VV} & vector-vector element-wise addition    \\
  6  & \texttt{SUB.VV} & vector-vector element-wise subtraction \\
  7  & \texttt{MUL.VS} & vector-scalar broadcast multiplication \\
  8  & \texttt{DIV.VS} & vector-scalar broadcast division       \\
  9  & \texttt{DOT.VV} & vector-vector dot product              \\
  10 & \texttt{NORM.V} & vector \(\ell_2\) norm                 \\
  \bottomrule
\end{tabular}
\end{center}
\vskip -0.1in
\end{table}

\subsection{Search Method}
\label{ssec:search_method}

Searching for programs is a combinatorial optimization problem, because the search space is finite when ignoring the initial values of hyperparameters. In this work, we develop a random search based method combined with several pruning techniques and intuitive priors. We choose random search due to several reasons. First, from a theoretical perspective, the no free lunch theorems for optimization \cite{Wolpert:1997e6} imply that random search is on average a good baseline method for combinatorial optimization. For example, random search based methods are shown to be competitive baselines in NAS \cite{Li:2019b2}. Second, from a practical perspective, random search is much simpler to implement efficiently and correctly than other methods, e.g., evolutionary search, and it can be surprisingly effective \cite{Balog:2017ba} when combined with other techniques, e.g., pruning techniques. Finally, random search can run in parallel by its nature, which helps us easily distribute tasks to multiple machines. For the hyperparameters, since the framework would adjust them automatically during the random walk process, we initialize them to a given fixed value to reduce implementation complexity and computational cost.

Unlike NAS works, in which the search spaces are usually constructed from expert-designed layers such that good architectures are dense in them, \sysname{}'s search space is constructed from a generic DSL such that good programs should be quite sparse. Naive random search would waste most computation on meaningless programs. We mitigate this issue by introducing four techniques specialized for the decision-based attack problem from two aspects --- the random program generating process and the search process. We will conduct an ablation study on these four techniques to show their effectiveness in Section~\ref{ssec:search_method_ablation_study}. 
%We explain them in detail in the following two paragraphs.

For the random program generating process, we apply two intuitive priors to improve the quality of the generated programs: (1) \emph{Compact program}: We use a program generating algorithm that prefers to generate programs with less unused operations. It is noted that our algorithm could still generate programs with many unused operations, but with a lower probability.
%which will encourage the program to explore the searching space more comprehensively. 
(2) \emph{Predefined operations}: We add three predefined operations \(\bm{v} = \bm{x}_0 - \bm{x}\), \(d = \| \bm{v} \|_2\), and \(\bm{u} = \bm{v} / d\) to the program before randomly generating the remaining operations. These predefined operations are common for decision-based attacks under the \(\ell_2\) norm, because the program needs to minimize the distance between \(\bm{x}_0\) and \(\bm{x}\). Thus the distance \(d\) between \(\bm{x}\) and \(\bm{x}_0\) and the direction \(\bm{u}\) from \(\bm{x}\) to \(\bm{x}_0\) should be useful. These operations all appear at the very beginning of many existing methods, including the Boundary attack \cite{Brendel:201866}, the Evolutionary attack~\cite{Dong:2019ae}, and the state-of-the-art Sign-OPT attack~\cite{Cheng:20200c}. Again, programs left these predefined operations unused could still be generated, but with a lower probability.
%\yinpeng{you mentioned probability, do we verify the probability is lower or any justification?}
Without reducing the size of our search space, both techniques just add priors to the generating process and increase the probability of generating better programs.

For the search process, we apply two pruning techniques to filter out trivially meaningless programs based on constraints imposed by the decision-based attack problem and the random walk algorithm template, including: (1) \emph{Inputs check}: We filter out programs that do not make use of all inputs, because they would be meaningless for the decision-based attack problem. This property is checked formally with some basic static analysis techniques \cite{Aho:19866d}. (2) \emph{Distance test}: We filter out programs that generate \(\bm{x}'\) violating the inequality \(\| \bm{x}' - \bm{x}_0 \|_2 < d_{\min}\) required by the framework in Alg.~\ref{alg:framework}. However, formally checking this property is extremely hard. Instead, we test this property on ten different inputs and filter out programs that fail in any of these tests. This informal test does not guarantee the inequality to hold for all inputs, but it works well in practice. The \emph{inputs check} and the \emph{distance test} are both done on CPU cores. By filtering out bad programs before they reach GPU, much less programs need to be evaluated on GPU. We will show that they save lots of expensive GPU computational cost for us in Section~\ref{ssec:searching_for_programs}.

\subsection{Evaluation Method}
\label{ssec:evaluation_method}

The last step is to define evaluation metrics such that we can distinguish good programs from bad ones. When evaluating the performance of decision-based attacks, we usually run them against many large deep models on different datasets to generate adversarial example for each sample in the test set. However, as running large models and attacking all samples in the test set are computationally expensive, this kind of evaluation is time-consuming and impractical for our problem with a huge search space. To address this issue, we leverage two strategies to make the evaluation fast and cheap.
%modified version
First, we adopt a shrunk by a factor of 0.5 version of EfficientNet-B0 \cite{Tan:2019ad} for evaluation.
%First, we adopt a slightly modified version of EfficientNet\footnote{We shrunk the EfficientNet-B0, the smallest variant of EfficientNets, by a factor of 0.5 to make it run even faster.} \cite{Tan:2019ad} for evaluation.
%\yinpeng{EfficientNets are DNNs, what's the meaning of widely used layers in DNNs}
EfficientNets are small and fast deep models that achieve high accuracies on various benchmarks. We train different binary classifiers for each pair of labels on the CIFAR-10 dataset. These classifiers can process more than 60,000 images per second on a single GTX 1080 Ti GPU. Second, we evaluate programs on a handful of examples and take an average over the evaluation metrics to save GPU time. Instead of using an absolute \(\ell_2\) distance between the original example \(\bm{x}_0\) and the best adversarial example \(\bm{x}\) the program found, we use a relative distance \(\| \bm{x} - \bm{x}_0 \|_2 / \| \bm{x}_1 - \bm{x}_0 \|_2\) as the metric where \(\bm{x}_1\) is the adversarial starting point as in Alg.~\ref{alg:framework}. We call it \emph{\(\ell_2\) distortion ratio}. A lower \(\ell_2\) distortion ratio means a better program.

% \junz{the connection to the above paragraph is not explicit: are they in parallel or progressive? Need some connecting words to make the logic coherent and smooth...}

Even with the small and fast classifier, running programs for tremendous random walk iterations is still unbearable computationally expensive. However, adopting lots of iterations is necessary for hyperparameters adjustment strategies to take effect in existing methods \cite{Brendel:201866,Dong:2019ae}.
To mitigate this issue, we first evaluate programs for a small number of iterations and select several top performing programs according to the evaluation metric.
% in previous small evaluation to evaluate for large number of iterations.
%\yinpeng{do not understand this sentence}
Then we perform a second round of evaluation of these programs for a larger number of iterations.
This evaluation strategy would also prefer choosing programs that achieve relatively high query-efficiency within few iterations. At the initial stage of the random walk process, the success rate is usually high, and thus the hyperparameters adjustment strategy tends to overshoot and harm the performance. So we disable hyperparameters adjustment in the small evaluation.

% \junz{not explicit connection, similar as above...}

We generate random programs in the SSA form as described in Section~\ref{ssec:search_space} and Section~\ref{ssec:search_method}.
Though SSA form programs are easy to analyze, they are slow and memory-inefficient to run. To make our SSA form programs run faster and occupy less memory, we compile them into their equivalent TAC form programs. During the compilation, we discard unused operations and allocate memory slots efficiently, such that the output TAC form programs are usually shorter and thus run faster with less memory usage than the original SSA form programs.

\section{Experiments}
\label{experiments}

In this section, we first run \sysname{} to search for top performing programs under the evaluation metric on the small classifiers as described in Section~\ref{ssec:evaluation_method}. We then compare the discovered algorithms with human designed attacks against different models on CIFAR-10 and ImageNet. Finally, we conduct an ablation study for the four techniques used in the search method of \sysname{} proposed in Section~\ref{ssec:search_method} to show their effectiveness.

\subsection{Searching for Programs}
\label{ssec:searching_for_programs}

%We search for top programs under the evaluation metric following the methods in Section~\ref{sec:methods}. 
We first introduce the detailed settings.
For the search space, we limit the maximum length of the program to 20 (i.e., the length of the Boundary attack's \texttt{generate()} function in \sysname{} DSL). We allow one scalar hyperparameter and set it to 0.01 initially.
We use the binary classifier for class 0 (airplane) and class 1 (automobile) of the CIFAR-10 dataset described in Section~\ref{ssec:evaluation_method}.
For the search method and the evaluation method,
we first generate programs randomly with all the four techniques introduced in Section~\ref{ssec:search_method}, then evaluate these programs in batches for 100 iterations to calculate their \(\ell_2\) distortion ratios. Each batch of programs is evaluated on one randomly selected example from the CIFAR-10 test set such that the \(\ell_2\) distortion ratios of these programs in the same batch can be compared with each other. The batch size here is 150, which achieves optimal performance on our hardware.
Second, we select the best program with lowest \(\ell_2\) distortion ratio from each batch of programs and evaluate it for 10,000 iterations on ten fixed examples from the CIFAR-10 test set to obtain their final \(\ell_2\) distortion ratios. Since the ten examples are fixed, these final \(\ell_2\) distortion ratios can be compared with each other.
% \yinpeng{How many batches are used? Why don't use the same 1 example across batches?}

We run this experiment for 50 times in parallel.
Each run allows a maximum number of 500 million queries to the classifier, which takes about two hours on one GTX 1080 Ti GPU.
In all 50 runs, we generate about 125 billion random programs. 45.475\% of these programs failed in the \emph{inputs check}, 54.497\% of them failed in the \emph{distance test}, and only 0.028\% of them survived both and continued to be evaluated against the classifier on GPU. These results show that the \emph{inputs check} and \emph{distance test} techniques save a lot of expensive GPU computational cost for us. As a result, we achieve a throughput of 294k programs per second per GTX 1080 Ti GPU.

\begin{figure*}
\begin{minipage}[t]{0.53\textwidth}
\begin{figure}[H]
\vskip 0.1in
\begin{center}
\includegraphics[width=0.44\linewidth,align=t,cfbox=lightgray 1pt 0.3em 0em]{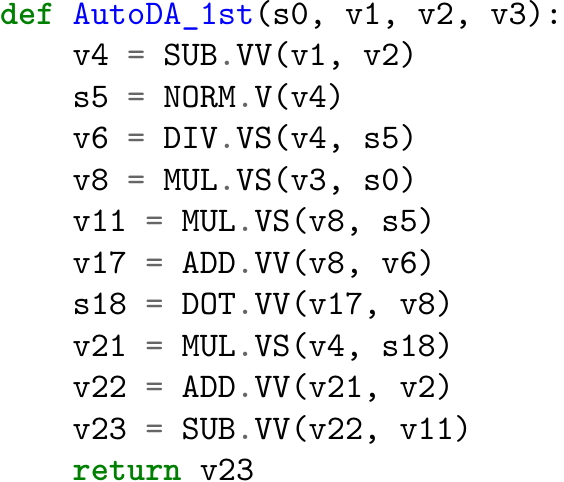}
\hspace{0.2em}
\includegraphics[width=0.44\linewidth,align=t,cfbox=lightgray 1pt 0.3em 0em]{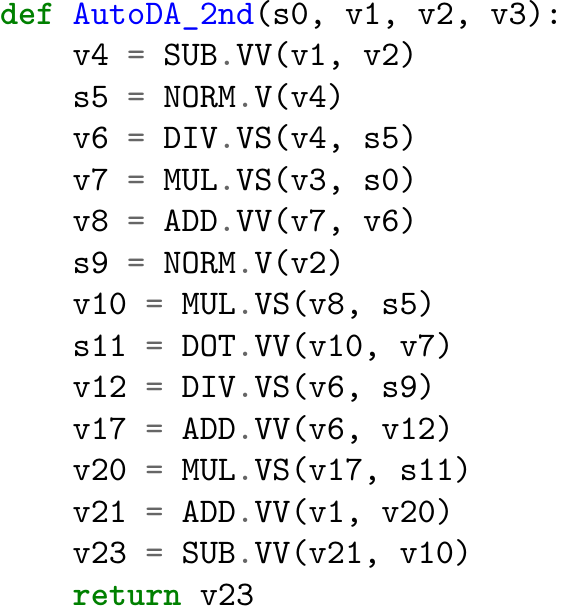}
\caption{The SSA form programs of \sysname{} 1st and 2nd, where \texttt{s0} is the hyperparameter, \texttt{v1} is the original example \(\bm{x}_0\), \texttt{v2} is the adversarial example \(\bm{x}\) the random walk process already found, and \texttt{v3} is the standard Gaussian noise \(\bm{n}\). The return value of these programs is the next random point to walk to. The \texttt{s}-prefix in variable's name means the variable is a scalar, and \texttt{v}-prefix for vector. Unused operations are discarded for clarity. The original programs as well as the compiled TAC form programs are provided in the supplementary material.}
\label{fig:programs}
\end{center}
\vskip -0.2in
\end{figure}
\end{minipage} \hspace{0.5em}
\begin{minipage}[t]{0.45\textwidth}
\begin{figure}[H]
\vskip 0.1in
\begin{center}
\centerline{\includegraphics[width=\linewidth]{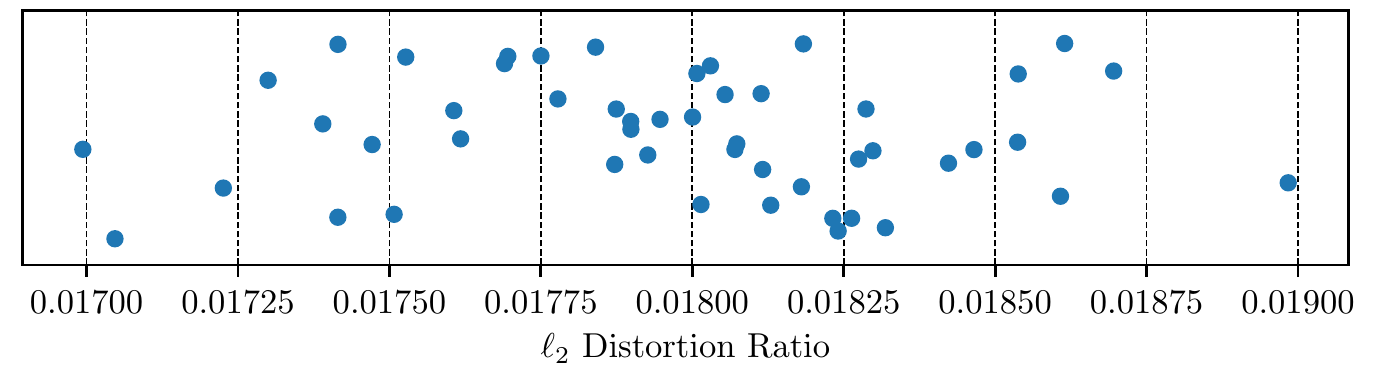}}
\vskip -0.1in
\caption{Distribution of the lowest \(\ell_2\) distortion ratio found in each of the 50 runs of searching for programs in our experiment.}
\label{fig:lra}
\end{center}
\vskip -0.3in
\end{figure}
\begin{figure}[H]
%\vskip 0.1in
\begin{center}
\centerline{\includegraphics[width=\linewidth]{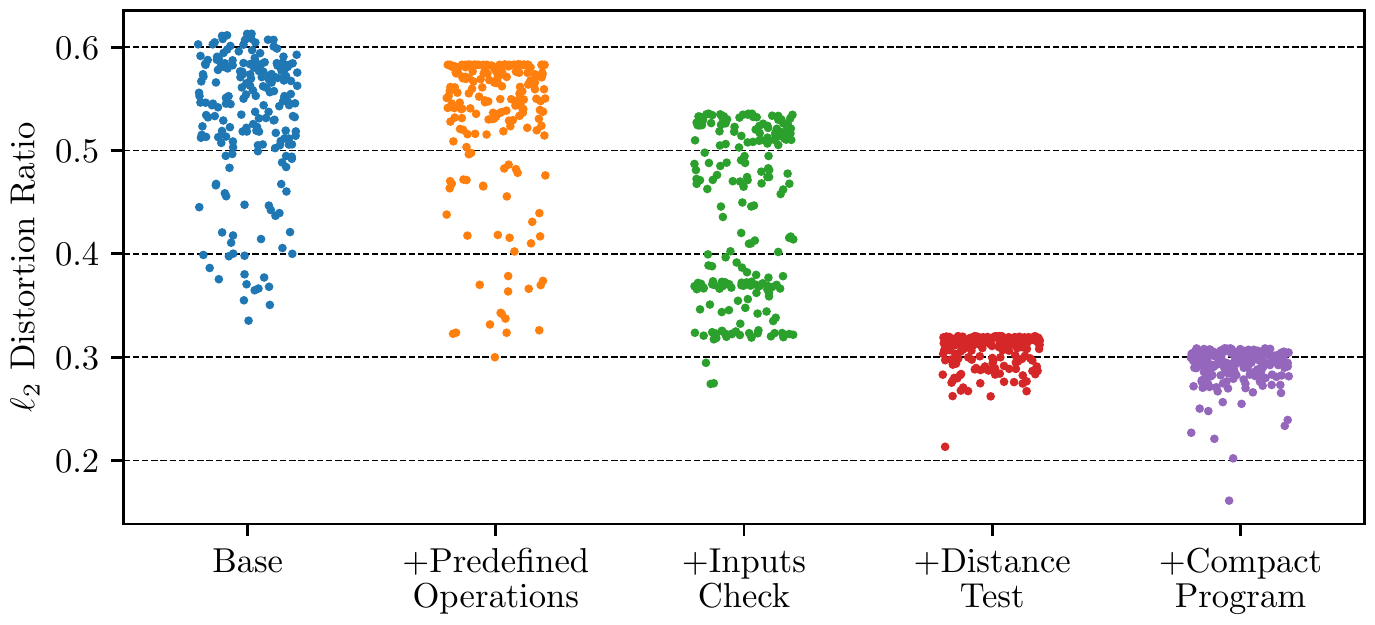}}
\vskip -0.1in
\caption{Search method ablation study experiment results. Each column shows the top 200 lowest \(\ell_2\) distortion ratios found by each search method. From left to right, each column adds a new technique.}
\label{fig:ablation}
\end{center}
\vskip -0.2in
\end{figure}
\end{minipage}
\end{figure*}

We plot the lowest \(\ell_2\) distortion ratio found on the ten fixed examples in each run in Figure~\ref{fig:lra}. They average at 0.01797 with a standard deviation of 0.00043.
% We plot the lowest \(\ell_2\) distortion ratio \yinpeng{w.r.t. what? 1 example with 100 iterations or 10 examples with 10000 iterations?} we found in each run in Figure~\ref{fig:lra}.
% Among the 50 runs, the best \(\ell_2\) distortion ratios on average is 0.01797 with a standard deviation of 0.00043.
The top one \(\ell_2\) distortion ratio is 0.01699, the second place is 0.01705, while the third place is 0.01723. The first place and the second place programs perform quite similarly and we choose both of them to compare with human designed attacks. We call them \emph{\sysname{} 1st} and \emph{\sysname{} 2nd}, respectively. We show the SSA form programs of \sysname{} 1st and 2nd in Figure~\ref{fig:programs}. We are surprised that they are quite short after discarding unused operations --- \sysname{} 1st only uses 10 operations and \sysname{} 2nd uses 13 operations, while the Boundary attack's \texttt{generate()} function uses 20 operations when expressed in the \sysname{} DSL. We also observe that \sysname{} 1st includes an operation sequence of \texttt{v8 = MUL(v3,s0)}; \texttt{s18 = DOT(v17,v8)}, and \sysname{} 2nd includes \texttt{v7 = MUL(v3,s0)}; \texttt{s11 = DOT(v10,v7)}, where \texttt{s0} is the scalar hyperparameter and \texttt{v3} is the Gaussian noise. They share a pattern of \mbox{\texttt{DOT(*,MUL(v3,s0))}}. A similar pattern is also observed in the Boundary attack \cite{Brendel:201866}. Moreover, both discovered attacks use the three predefined operations which are also used in existing works. These similarities qualitatively suggest the rationality of some heuristics used in existing attack methods.

\subsection{Results on CIFAR-10 and ImageNet}
\label{ssec:results_on_cifar_10_and_imagenet_models}

We benchmark the \sysname{} 1st and 2nd programs we found for attacking various models under the \(\ell_2\) norm untargeted decision-based threat model on the CIFAR-10 \cite{krizhevsky2009learning} and ImageNet \cite{Deng:2009ff} datasets, and compare them with existing methods. We follow \citeauthor{Dong:20200a}'s benchmark methodology: We consider one attack to be successful after it finds adversarial example whose \(l_2\) distance w.r.t. the original example is smaller than \(\epsilon = 1.0\) on CIFAR-10, and whose normalized \(\ell_2\) distance w.r.t. the original example is smaller than \(\epsilon = \sqrt{0.001}\) on ImageNet (normalized \(\ell_2\) distance is defined as \(\| \cdot \|_2 / \sqrt{d}\) where \(d\) is the dimension of the input to the classifier). Then we use the \emph{attack success rate vs. queries} curve to show the effectiveness and efficiency of these attack algorithms, as well as the \emph{\(\ell_2\) distortion vs. queries} curve widely used in previous decision-based attack works \cite{Brendel:201866,Cheng:20200c}.

\begin{figure}[t]
\vskip 0.1in
\begin{center}
\centerline{\includegraphics[width=\columnwidth]{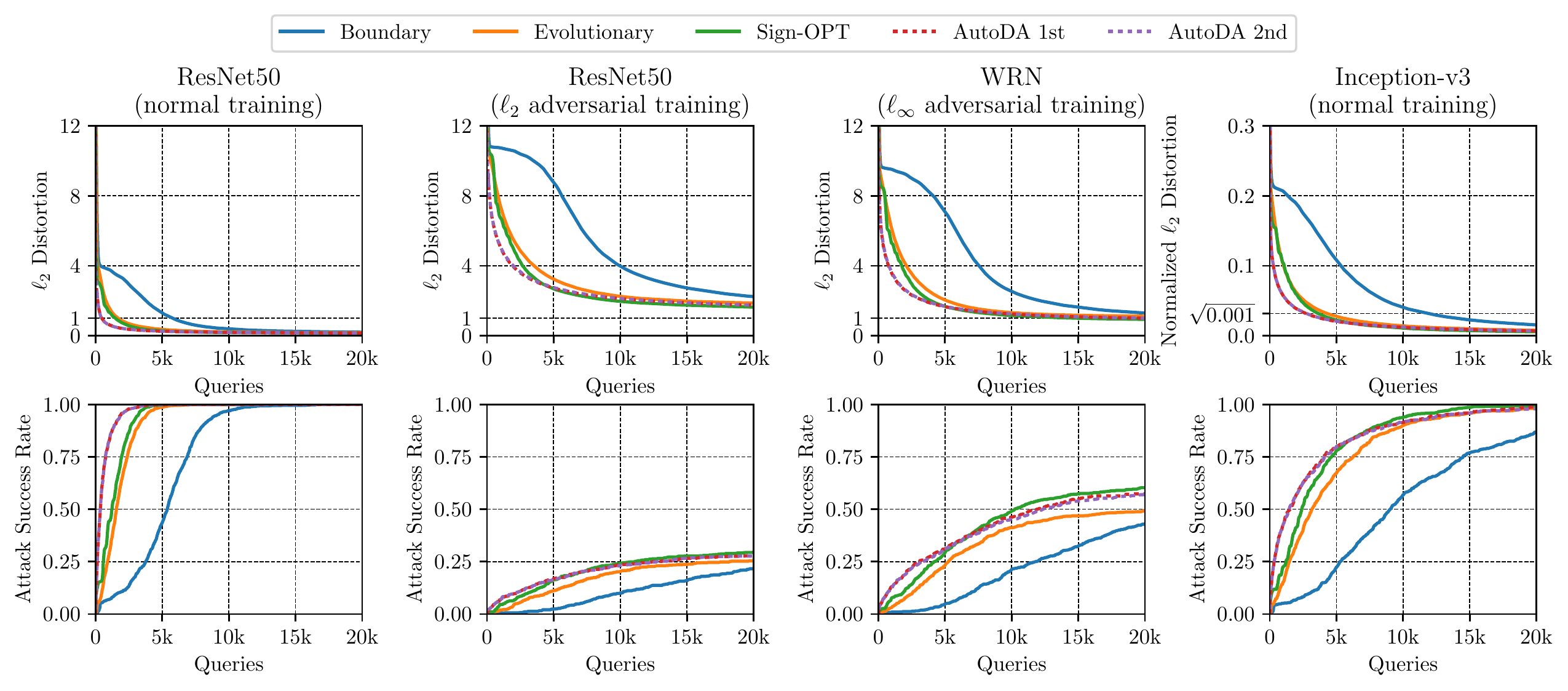}}
\caption{The \emph{\(\ell_2\) distortion vs. queries} and \emph{attack success rate vs. queries} curves on the three models on the CIFAR-10 dataset and the Inception-v3 model on the ImageNet dataset.}
\label{fig:baseline}
\end{center}
\vskip -0.2in
\end{figure}

We compare the following untargeted decision-based  attack methods with our \sysname{} 1st and 2nd: (1) The Boundary attack \cite{Brendel:201866}, (2) the Evolutionary attack \cite{Dong:2019ae}, and (3) the Sign-OPT attack \cite{Cheng:20200c}. The first two attacks are both based on the random walk framework. They are included in \citeauthor{Dong:20200a}'s benchmark, so we adapt their implementations. However, we disable the dimension reduction trick for these two attacks on ImageNet because we want to know the original attacks' strength. The third one is a recently proposed query-efficient attack based on zeroth-order optimization. We adapt the implementation from its official repository and leave all hyperparameters unmodified. 
Together with our \sysname{} 1st and 2nd attacks, we have five attacks to run. As for the initialization of hyperparameters in \sysname{} 1st and 2nd, we adopt the original value 0.01 used in the search to attack the CIFAR-10 models. However, when transferred to ImageNet, they would fail to pass the distance test with their hyperparameters initialized to 0.01. To overcome this issue caused by changing dimension, we decrease their hyperparameters' initial value to 0.001 on ImageNet. The Sign-OPT attack might spend up to several hundreds of queries for finding the starting points, while other attacks do not. Thus we include these queries for finding the starting points in the total queries to make the comparison more fair. More details on how we run the five attacks are provided in the supplementary material.

We select the first 1,000 images from the CIFAR-10 test set and the first 1,000 images from the ImageNet test set to run our benchmark. We choose the normally trained ResNet50 \cite{He:201604} model on the CIFAR-10 dataset and normally trained Inception-v3 \cite{Szegedy:2016bd} model on the ImageNet dataset both provided by torchvision \cite{Paszke:201942} for the five methods to attack. Besides, we also aim to understand the strength of our attacks on stronger models, and thus we include \(\ell_2\) adversarially trained (\(\epsilon = 1.0\)) ResNet50 model provided by \citet{robustness} and \(\ell_\infty\) adversarially trained (\(\epsilon = 8/255\)) WRN model \cite{Zagoruyko:20160e} provided by \citet{Madry:2018e9}. The clean accuracies for these models on our benchmark examples are: 94.4\% for the normally trained ResNet50 model, 78.1\% for the normally trained Inception-v3 model, 82.4\% for the \(\ell_2\) adversarially trained ResNet50 model, and 87.3\% for the \(\ell_\infty\) adversarially trained WRN model.

% \begin{figure}[t]
% \vskip 0.1in
% \begin{center}
% \centerline{\includegraphics[width=\columnwidth]{res/imagenet.pdf}}
% \caption{The \emph{\(\ell_2\) distortion vs. queries} and \emph{attack success rate vs. queries} curves on the Inception-v3 model on the ImageNet dataset.}
% \label{fig:imagenet}
% \end{center}
% \vskip -0.2in
% \end{figure}

We plot the \emph{\(\ell_2\) distortion vs. queries} and \emph{attack success rate vs. queries} curves for the five attacks on the three models on the CIFAR-10 dataset and the Inception-v3 model on the ImageNet dataset in Figure~\ref{fig:baseline}. We also provide \emph{attack success rate} at different number of \emph{queries} in Table~\ref{tab:asr} for numerical comparisons.

\begin{table*}[t]
\caption{The \emph{attack success rate} given different number of \emph{queries} on the three models on the CIFAR-10 dataset and the Inception-v3 model on the ImageNet dataset.}
\label{tab:asr}
\vskip 0.1in
\begin{center}
\begin{footnotesize}
\setlength{\tabcolsep}{1.9pt}
\begin{tabular}{c|rrr|rrr|rrr|rrr}
\toprule
Model        & \multicolumn{3}{c|}{\thead{ResNet50 \\ (normal training)}}
             & \multicolumn{3}{c|}{\thead{ResNet50 \\ (\(\ell_2\) adv. training)}}
             & \multicolumn{3}{c|}{\thead{WRN \\ (\(\ell_\infty\) adv. training)}}
             & \multicolumn{3}{c}{\thead{Inception-v3 \\ (normal training)}} \\
\midrule
Queries      & \multicolumn{1}{c}{2,000} & \multicolumn{1}{c}{4,000} 
             & \multicolumn{1}{c|}{20,000}
             & \multicolumn{1}{c}{2,000} & \multicolumn{1}{c}{4,000} 
             & \multicolumn{1}{c|}{20,000}
             & \multicolumn{1}{c}{2,000} & \multicolumn{1}{c}{4,000} 
             & \multicolumn{1}{c|}{20,000}
             & \multicolumn{1}{c}{2,000} & \multicolumn{1}{c}{4,000} 
             & \multicolumn{1}{c}{20,000} \\
\midrule \midrule
Boundary       & 10.7\% & 28.4\% & 100.0\% &  0.6\% &  1.6\% & 21.6\% &  1.1\% &  2.7\% & 43.1\% &  7.0\% & 15.0\% & 86.8\% \\
Evolutionary   & 64.9\% & 96.3\% & 100.0\% &  4.4\% &  8.9\% & 25.4\% &  7.7\% & 18.2\% & 49.4\% & 33.4\% & 59.2\% & 98.1\% \\
Sign-OPT       & 76.1\% & 98.8\% & 100.0\% &  6.6\% & 12.6\% & \textbf{29.5\%} & 11.3\% & 24.1\% & \textbf{60.3\%} & 41.4\% & 69.1\% & \textbf{99.2\%} \\
\midrule
\sysname{} 1st & \textbf{95.9\%} & \textbf{99.7\%} & 100.0\% &  9.7\% & \textbf{14.9\%} & 27.8\% & \textbf{19.2\%} & \textbf{28.3\%} & 57.4\% & \textbf{57.1\%} & \textbf{74.6\%} & 98.7\% \\
\sysname{} 2nd & 95.6\% & 99.5\% & 100.0\% & \textbf{10.0\%} & 14.8\% & 27.7\% & 18.9\% & 27.1\% & 57.0\% & 56.5\% & 73.4\% & 97.8\% \\
\bottomrule
\end{tabular}
\end{footnotesize}
\end{center}
\vskip -0.1in
\end{table*}

From these curves and the table, we can observe that \sysname{} 1st performs slightly better than \sysname{} 2nd, which is consistent with their quite-close \(\ell_2\) distortion ratios shown in Section~\ref{ssec:searching_for_programs}. This fact suggests the rationality of using the \(\ell_2\) distortion ratio as the evaluation metric. Moreover, both the \sysname{} 1st and 2st outperform the other three human designed baselines by a lot when the number of queries is smaller than 5,000. When the number of queries grows larger than 5,000, our \sysname{} 1st attack method still outperforms the Boundary attack and the Evolutionary attack, while becomes slightly behind the Sign-OPT attack method, and only for defensive models and ImageNet model, this small gap is noticeable. These behaviors are consistent on all models and datasets we run our experiments on, demonstrating the great query-efficiency of our discovered attack methods especially under low number of queries, which is important in real-world scenarios for black-box attacks \cite{Brendel:201866,Ilyas:20185d}.

\subsection{Ablation Study on Search Method}
\label{ssec:search_method_ablation_study}

As described in Section~\ref{ssec:search_method}, we apply four techniques to our search method, which are (1) \emph{Predefined operations}, (2) \emph{Inputs check}, (3) \emph{Distance test}, and (4) \emph{Compact program}. To illustrate their effectiveness, we conduct the following ablation study on search method. Starting from the base search method using only naive random search, we add the four techniques one by one, so we get five different random search methods including the base one. For each of these five random search methods, we run it to evaluate 100,000 programs against the classifier for 100 iterations on five fixed examples and calculate the \(\ell_2\) distortion ratio for each program. We plot the top 200 lowest \(\ell_2\) distortion ratios that each search method found in Figure~\ref{fig:ablation}.

From the figure, we can observe that with more techniques added, the top 200 lowest \(\ell_2\) distortion ratios overall show a decreasing trend. For example, the lowest \(\ell_2\) distortion ratio in each column becomes lower and lower. These results demonstrate the effectiveness of the four techniques we applied to our search method. These results are qualitative, because these techniques might interfere with each other so that multiple techniques combined might bring improvement larger than the sum of improvements brought by applying each of them.
As a result, the absolute improvement shown in the figure does not imply the effectiveness of each technique.

\section{Conclusion and Discussion}
\label{sec:conclusion_and_discussion}

In this work, we propose to automate the process of discovering decision-based attack algorithms. Starting from the random walk framework as the algorithm template, we construct our generic search space from the \sysname{} DSL, explore this search space using random search integrated with several pruning techniques and intuitive priors, and evaluate programs in the search space using a small and fast model. The discovered attack algorithms are simple, while consistently achieve high query-efficiency when transferred to both normal and defensive models on the CIFAR-10 and ImageNet datasets.

Many future extensions can be done to this work. First, we particularly focus on the untargeted decision-based threat model under the \(\ell_2\) norm in this work. Extending our approach to targeted attack should be straightforward, while extending to the \(\ell_\infty\) norm might need more efforts, because designing another search space specialized for the \(\ell_\infty\) norm is necessary. Second, we limit the search space to be relatively small and use a random search based method to explore it. More advanced search methods like evolutionary search and more computational resources could explore larger and more powerful search space, which should lead to better algorithms. Finally, advanced static analysis tools can help us simplify the discovered attack algorithms and identify important operations in these algorithms.
%, and understand how these algorithms work.

% \begin{ack}
% \end{ack}

%\section*{References}

\bibliography{neurips_2021}

\clearpage
\appendix

\section{Hyperparameters Adjustment Strategies}
\label{sec:hyperparameters_adjustment}

In this section, we will introduce the hyperparameters adjustment strategies used in the Boundary attack \cite{Brendel:201866}, the Evolutionary attack \cite{Dong:2019ae}, and our \sysname{}.

\textbf{Boundary attack.} The Boundary attack \cite{Brendel:201866} has two hyperparameters --- the total perturbation \(\delta\) and the length of the step \(\epsilon\) towards the original input. The original implementation would adjust both hyperparameters during the random walk process. If the success rate for the past several trails is too low, \(\epsilon\) would be decreased and vise versa. The attack is converged when \(\epsilon\) reaches zero. The \(\delta\) would be adjusted according to the so called orthogonal perturbation's success rate similar to adjusting \(\epsilon\). However, using a fixed \(\delta\) during the random walk process has negligible performance impact with a proper initial value. Under this case, we only need to adjust \(\epsilon\) during the random walk process.

\textbf{Evolutionary attack.} The Evolutionary attack \cite{Dong:2019ae} has several hyperparameters, while only the \(\mu\) is adjusted during the random walk process. The \(\mu\) is adjusted once for every \(T\) iterations as follows:
\begin{equation}
  \mu \leftarrow \mu \cdot \exp{(p - \bar{p})}
\end{equation}
where \(p\) is the success rate of past \(T\) trails, \(\bar{p}\) is a predefined target success rate. This is a negative feedback strategy keeping the success rate around \(\bar{p}\). The original implementation set \(T\) to 30 and \(\bar{p}\) to 0.2.

\textbf{\sysname{}.} Our \sysname{} uses a negative feedback strategy similar to the ones used in the above two attack methods especially the Evolutionary attack. Instead of using the \(\exp(p-\bar{p})\) function in the Evolutionary attack, we use a piecewise linear function \(f\) that satisfies \(f(0)=l\), \(f(1)=h\), and \(f(\bar{p})=1\), where \(l,h\) are both predefined constants satisfying \(0<l<1<h\). The \(\bar{p}\) is the predefined target success rate same as in the Evolutionary attack. For implementation simplicity, instead of adjusting the hyperparameter every \(T\) iterations as in the Evolutionary attack, we adjust the hyperparameter in each iteration according to a decayed success rate \(p\) defined as follows:
\begin{equation}
  p \leftarrow \alpha \cdot p + (1-\alpha) \cdot k
\end{equation}
where \(k = 1\) if \(\bm{x}'\) is adversarial, otherwise \(k = 0\), \(\alpha\) is the decay rate. We adjust the hyperparameter \(s\) as follows:
\begin{equation}
  s \leftarrow s \cdot \left[ f(p) \right]^{1/10}
\end{equation}
where the extra \(\cdot^{1/10}\) is to stabilize the hyperparameter adjustment, so that when \(p\) is quite close to 0.0 for ten iterations, the \(s\) would be decreased to at most \(l \cdot s\) instead of a much smaller \(l^{10} \cdot s\), and when \(p\) is quite close to 1.0 for ten iterations, the \(s\) would be increased to at most \(h \cdot s\) instead of a much larger \(h^{10} \cdot s\). This negative feedback strategy would keep the success rate around \(\bar{p}\), too. We set the decay rate \(\alpha\) to 0.95, \(l\) to 0.5, \(h\) to 1.5, and \(\bar{p}\) to 0.25 in \sysname{}.

\section{The \sysname{} 1st and \sysname{} 2nd}

\begin{figure*}[ht]
\vskip 0.1in
\begin{center}
\subfigure[\raisebox{1em}{}]{%
  \begin{tikzpicture}
  \node[anchor=north west, inner sep=0pt] (1st_full) at (0em, 0em)
    {\includegraphics[height=15.8em,keepaspectratio,align=t,cfbox=lightgray 1pt 0.3em 0em]{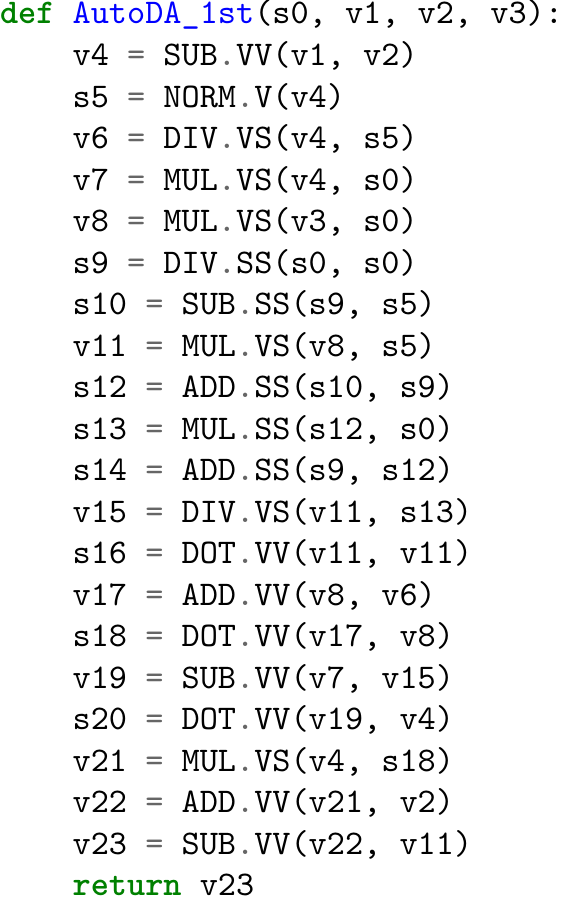}};
  \path let \p1 = (1st_full.north east)
    in node[anchor=north west, inner sep=0pt] (1st) at (\x1+0.5em,\y1)
    {\includegraphics[height=6.9em,keepaspectratio,align=t,cfbox=lightgray 1pt 0.3em 0em]{res/AutoDA_1st.pdf}};
  \path let \p1 = (1st_full.south east)
    in node[anchor=south west, inner sep=0pt] (1st_tac) at (\x1+0.5em,\y1)
    {\includegraphics[height=6.9em,keepaspectratio,align=t,cfbox=lightgray 1pt 0.3em 0em]{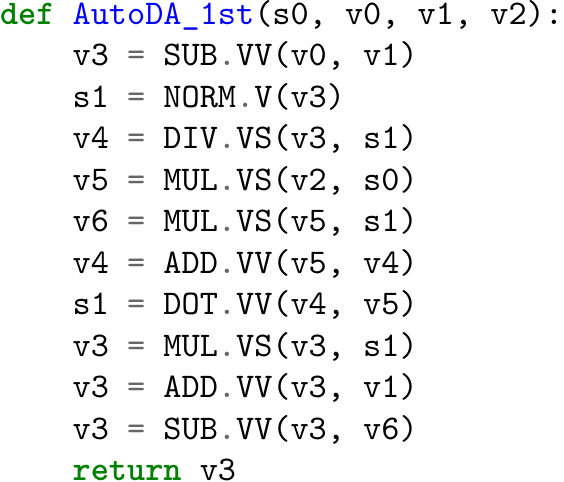}};
  \draw [draw=gray, >=latex, ->, semithick, rounded corners=5pt]
    (1st_full.north) -- ++(0,1.8em) -| (1st.north)
    node[pos=0.25,below] {\footnotesize{discard unused operations}};
  \draw [draw=gray, >=latex, ->, semithick, rounded corners=5pt]
    (1st.south) -- (1st_tac.north)
    node[pos=0.5] {\footnotesize{allocate memory slots}};
  \end{tikzpicture}\label{sfig:1st}%
}
\subfigure[\raisebox{1em}{}]{%
  \begin{tikzpicture}
  \node[anchor=north west, inner sep=0pt] (2nd_full) at (0em, 0em)
    {\includegraphics[height=15.8em,keepaspectratio,align=t,cfbox=lightgray 1pt 0.3em 0em]{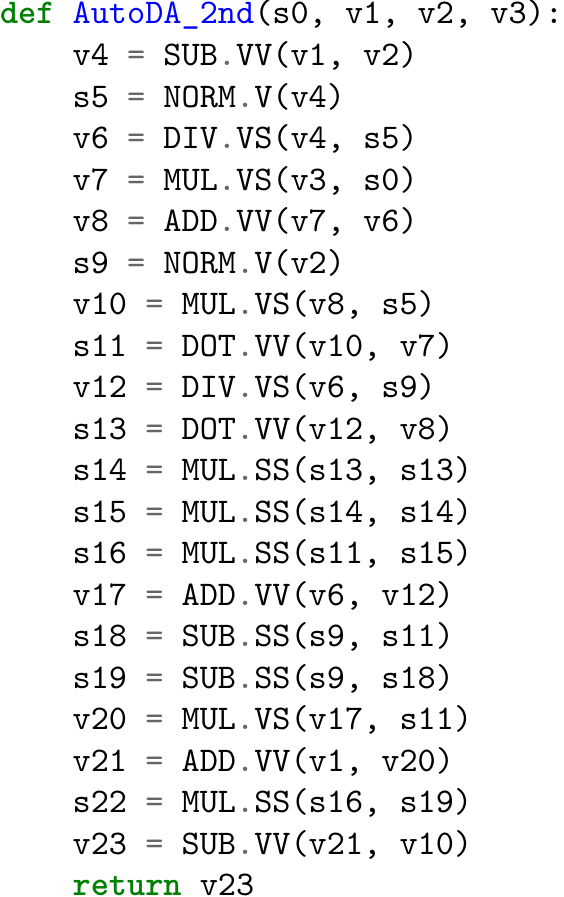}};
  \path let \p1 = (2nd_full.north east)
    in node[anchor=north west, inner sep=0pt] (2nd) at (\x1+0.5em,\y1)
    {\includegraphics[height=6.9em,keepaspectratio,align=t,cfbox=lightgray 1pt 0.3em 0em]{res/AutoDA_2nd.pdf}};
  \path let \p1 = (2nd_full.south east)
    in node[anchor=south west, inner sep=0pt] (2nd_tac) at (\x1+0.5em,\y1)
    {\includegraphics[height=6.9em,keepaspectratio,align=t,cfbox=lightgray 1pt 0.3em 0em]{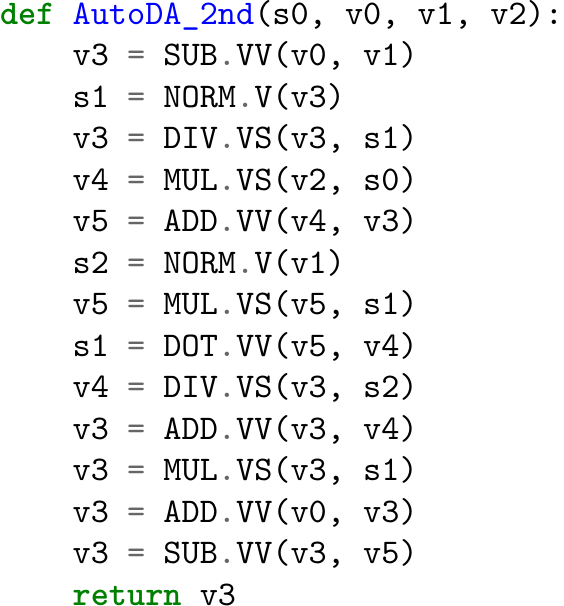}};
  \draw [draw=gray, >=latex, ->, semithick, rounded corners=5pt]
    (2nd_full.north) -- ++(0,1.8em) -| (2nd.north)
    node[pos=0.25,below] {\footnotesize{discard unused operations}};
  \draw [draw=gray, >=latex, ->, semithick, rounded corners=5pt]
    (2nd.south) -- (2nd_tac.north)
    node[pos=0.5] {\footnotesize{allocate memory slots}};
  \end{tikzpicture}\label{sfig:2nd}%
}
\caption{\protect\subref{sfig:1st} and \protect\subref{sfig:2nd} show the original SSA form programs, the SSA form programs after discarding unused operations, and the compiled TAC form programs after allocating memory slots for the SSA form programs, for the \sysname{} 1st and \sysname{} 2nd respectively. In the original SSA form programs and the SSA form programs after discarding unused operations, \texttt{s0} is the hyperparameter, \texttt{v1} is the original adversarial example \(\bm{x_0}\), \texttt{v2} is the adversarial example \(\bm{x}\) the random walk process already found, and \texttt{v3} is the standard Gaussian noise \(\bm{n}\). In the TAC form programs, \texttt{s0}, \texttt{v0}, \texttt{v1} and \texttt{v2} is compiled from the \texttt{s0}, \texttt{v1}, \texttt{v2}, \texttt{v3} in the SSA form programs respectively.}
\label{fig:detailed_programs}
\end{center}
\end{figure*}

For the \sysname{} 1st and \sysname{} 2nd, we show their original SSA form programs, their SSA form programs after discarding unused operations, and their compiled TAC form programs in Figure~\ref{fig:detailed_programs}.

\section{Additional Details on the \sysname{} DSL}

\begin{figure}[htbp]
% \vskip 0.1in
\begin{center}
\subfigure[\raisebox{1em}{}]{\includegraphics[width=0.25\columnwidth,align=t,cfbox=lightgray 1pt 0.3em 0em]{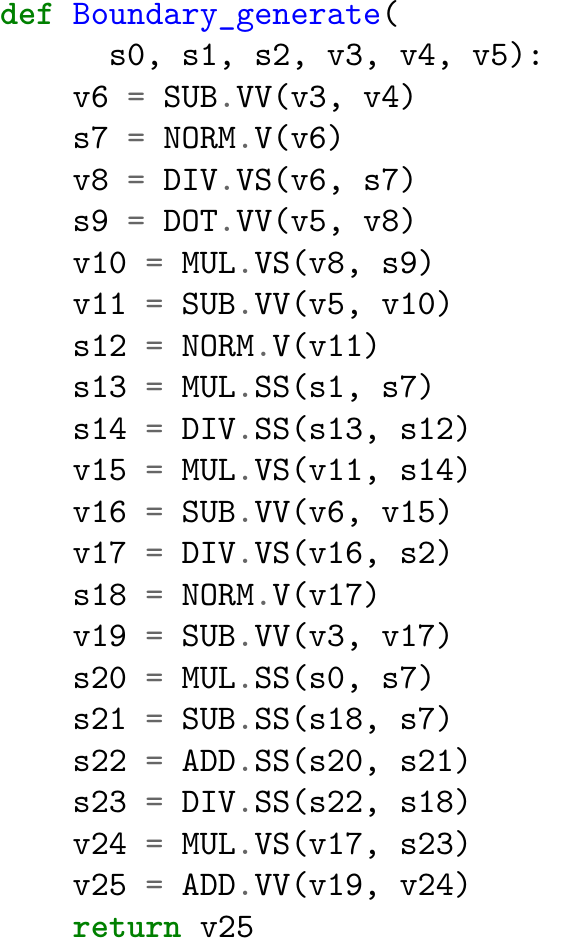}\label{sfig:boundary}}
\hspace{0.2em}
\subfigure[\raisebox{1em}{}]{\includegraphics[width=0.25\columnwidth,align=t,cfbox=lightgray 1pt 0.3em 0em]{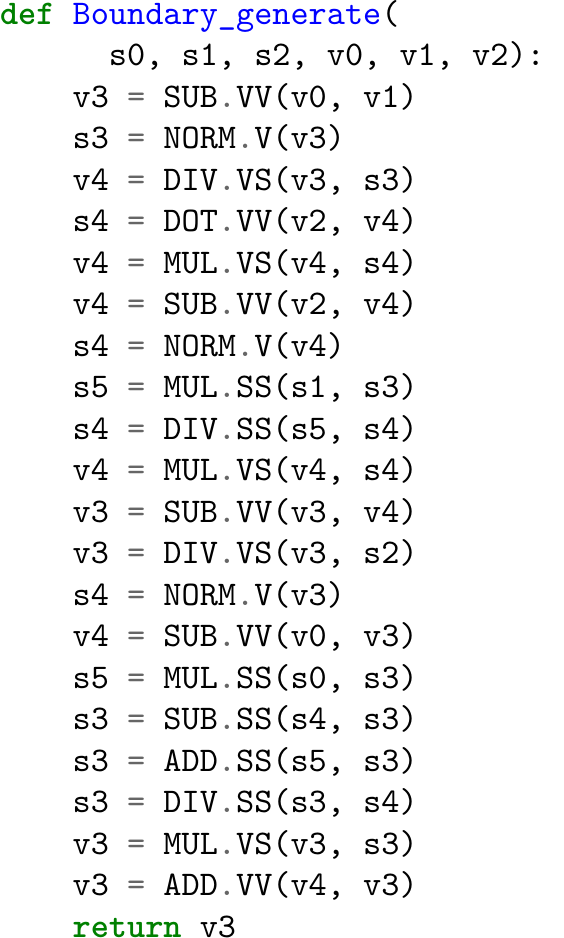}\label{sfig:boundary_tac}}
\caption{\protect\subref{sfig:boundary} One possible implementation of the Boundary attack's \texttt{generate()} function as a SSA form program in the \sysname{} DSL. \texttt{s0} and \texttt{s1} are the \(\epsilon\) and \(\delta\) hyperparameters in the Boundary attack respectively, \texttt{s2} is derived from the \(\delta\) as \(\sqrt{1+\delta^2}\). The Boundary attack would adjust both the \(\epsilon\) and the \(\delta\) during the random walk process. However, a fixed \(\delta\) has negligible performance impact as mentioned in Section~\ref{sec:hyperparameters_adjustment}. For simplicity, we fix the \(\delta\) so that both \texttt{s1} and \texttt{s2} can be considered as fixed hyperparameters. As a result, we do not need to add the extra \(\sqrt{\cdot}\) operation to our DSL for implementing the Boundary attack. \texttt{v3} is the original example \(\bm{x}_0\), \texttt{v4} is the adversarial example \(\bm{x}\) the random walk process already found, and \texttt{v5} is the standard Gaussian noise \(\bm{n}\). \protect\subref{sfig:boundary_tac} The compiled TAC form version of the same program. The \texttt{s0}, \texttt{s1}, \texttt{s2}, \texttt{v0}, \texttt{v1} and \texttt{v2} are compiled from the \texttt{s0}, \texttt{s1}, \texttt{s2}, \texttt{v3}, \texttt{v4} and \texttt{v5} in the SSA form program respectively.}
\label{fig:boundary}
\end{center}
\vskip -0.2in
\end{figure}

We provide the detailed definitions of available operations in the \sysname{} domain specific language (DSL) in Table~\ref{tab:detailed_ops_list}. To show the expressiveness of our DSL, we provide one possible implementation of the Boundary attack \cite{Brendel:201866} using our \sysname{} DSL in Figure~\ref{fig:boundary}.

\begin{table*}[ht]
\caption{List of available operations in the \sysname{} DSL. The suffix of each operation's notation indicates the parameters' type of the operation, where \texttt{S} denotes scalar type, and \texttt{V} denotes vector type. For example, the \texttt{VS} suffix means the operation's first parameter is a scalar and second parameter is a vector. In the parameter(s) column, \(a\) or \(\vec{a}\) stands for the first parameter, \(a\) for scalar and \(\vec{a}\) for vector; \(b\) or \(\vec{b}\) stands for the second parameter, \(b\) for scalar and \(\vec{b}\) for vector. In the output column, \(r\) for scalar output, and \(\vec{r}\) for vector output. In the mathematical expression column, the subscript \(\cdot_i\) on vector variable means the \(i\)-th component of the vector, \(\forall i\) means for all dimension of the vector.}
\label{tab:detailed_ops_list}
\vskip 0.1in
\begin{center}
\begin{small}
\setlength{\tabcolsep}{4.5pt}
\renewcommand{\arraystretch}{1.2}
\begin{tabular}{lllccl}
  \toprule
  ID & Notation        & Description & Parameter(s) & Output & Mathematical Expression \\
  \midrule
  1  & \texttt{ADD.SS} & scalar-scalar addition & \(a,b\) & \(r\) & \(r = a + b\) \\
  2  & \texttt{SUB.SS} & scalar-scalar subtraction & \(a,b\) & \(r\) & \(r = a - b\) \\
  3  & \texttt{MUL.SS} & scalar-scalar multiplication & \(a,b\) & \(r\) & \(r = a b\) \\
  4  & \texttt{DIV.SS} & scalar-scalar division & \(a,b\) & \(r\) & \(r = a / b\) \\
  5  & \texttt{ADD.VV} & vector-vector element-wise addition & \(\vec{a},\vec{b}\) & \(\vec{r}\) & \(\vec{r} = \vec{a} + \vec{b}\) \\
  6  & \texttt{SUB.VV} & vector-vector element-wise subtraction & \(\vec{a},\vec{b}\) & \(\vec{r}\) & \(\vec{r} = \vec{a} - \vec{b}\) \\
  7  & \texttt{MUL.VS} & vector-scalar broadcast multiplication & \(\vec{a},b\) & \(\vec{r}\) & \(r_i = a_i b, \forall i\) \\
  8  & \texttt{DIV.VS} & vector-scalar broadcast division & \(\vec{a},b\) & \(\vec{r}\) & \(r_i = a_i / b, \forall i\) \\
  9  & \texttt{DOT.VV} & vector-vector dot product & \(\vec{a},\vec{b}\) & \(r\) & \(r = \vec{a} \cdot \vec{b} \) \\
  10 & \texttt{NORM.V} & vector \(\ell_2\) norm & \(\vec{a}\) & \(r\) & \(r = \| \vec{a} \|_2\) \\
  \bottomrule
\end{tabular}
\end{small}
\end{center}
% \vskip -0.1in
\end{table*}

\section{Additional Implementation Details}

We need to implement the \sysname{} system efficiently, because our task is computational intensive even with the computational cost reducing techniques described in Section~3. As a result, we implement the \sysname{} system in the C++ programming language. We train our binary classifiers in Python using a keras \cite{chollet2015keras} implementation~\footnote{\url{https://github.com/qubvel/efficientnet}} of EfficientNet \cite{Tan:2019ad}, then export them to a proper format, so that we can run these classifiers using TensorFlow \cite{tensorflow2015-whitepaper} for C. These classifiers require large batch size ($>1,000$) to achieve their full speed, so that we have to run a large number of programs in parallel to fulfill this requirement. As a result, we divide these large number of programs into smaller batches (e.g.,
with batch size of 150 as mentioned in Section~4.1), and execute these smaller batches in parallel on multiple threads to utilize multiple CPU cores. We also use the Eigen C++ template library for linear algebra \cite{eigenweb} to execute programs on CPU for better performance.
 with batch size of 150 as mentioned in Section~4.1), and execute these smaller batches in parallel on multiple threads to utilize multiple CPU cores. We also use the Eigen C++ template library for linear algebra \cite{eigenweb} to execute programs on CPU for better performance.
We will provide additional implementation details on random program generating and the SSA form to TAC form compiler for the rest of this section.

\textbf{Random program generating.} We describe in detail how we generate a random program starting from an empty program in this paragraph: (1) Add the three \emph{predefined operations} described in Section~3.3 to the empty program. (2) Keep randomly and uniformly selecting one operation from all available operations in the \sysname{} DSL and append it to the program. We keep a record of unused operations, so that when randomly choosing the parameter(s) of a new operation, we choose unused operation's outputs with higher probability. This is the \emph{compact program} technique mentioned in Section~3.3. (3) For the last operation before the program reaching maximum length, we consider it as the output of the program, so that this operation must output a vector. As a result, we randomly select a vector output operation as the last operation. After generating each random program, we run the \emph{inputs check} and the \emph{distance test} as described in Section~3.3. The whole process of generating random programs then running the \emph{inputs check} and the \emph{distance test} is CPU intensive. To mitigate this issue, we do random program generating, checking and testing in parallel on multiple threads.

\textbf{SSA form to TAC form compiler.} As mentioned in Section~3.4, we compile SSA form programs to their equivalent TAC form programs before executing them for better performance and smaller memory usage. This simple SSA form to TAC form compiler first discards unused operations then allocates memory slots to get the equivalent TAC form program. Discarding unused operations can be solved with some simple compiler techniques, while allocating memory slots with optimal memory usage is NP-complete \cite{Aho:19866d}. Instead, we use a non-optimal linear complexity algorithm to allocate memory slots, which is fast and produces TAC form programs with reasonable quality. For example, for the \sysname{} 1st shown in Figure~\ref{sfig:1st}, the SSA form program after discarding unused operations still use three scalar memory slots and eleven vector memory slots, while the compiled TAC form program use two scalar memory slots and seven vector memory slots. For the \sysname{} 2nd shown in Figure~\ref{sfig:2nd} and the Boundary attack's \texttt{generate()} function shown in Figure~\ref{fig:boundary}, we can observe similar memory usage reduction.

\section{Detailed Experiment Setups}

We describe in detail how we run our \sysname{} 1st, \sysname{} 2nd, the Boundary attack, the Evolutionary attack, and the Sign-OPT attack in this paragraph. We adapt the implementations of the Boundary attack and the Evolutionary attack from \citeauthor{Dong:20200a}'s benchmark~\footnote{\url{https://github.com/thu-ml/ares}}. We adapt the implementation of the Sign-OPT attack from its official repository~\footnote{\url{https://github.com/cmhcbb/attackbox}}. We remove the dimension reduction technique in the Boundary attack and the Evolutionary attack's implementations, which is a useful technique for accelerating black-box attacks \cite{chen2017zoo} when attacking models on ImageNet, because we want to know the original attacks' strength. Though this technique can be applied to our \sysname{} 1st and 2nd easily, the Sign-OPT attack's implementation does not support it, so disabling it in all attacks also makes the comparison more fair. For all attacks, we clip the inputs into \([0,1]\) to make them valid images before running the classifier for prediction labels. The Sign-OPT attack's original implementation would fail to find starting points for some inputs. We modify its implementation to fallback to starting points selected from the test set after 100 failures. For our \sysname{} 1st and 2nd as well as the Boundary attack and the Evolutionary attack, we find starting points by keeping adding different standard Gaussian noises to the original examples until finding one that causes misclassification, and fallback to starting points selected from the test set after 100 failures similar to the Sign-OPT attack.

\end{document}